\documentclass[final]{cvpr}

\usepackage{times}
\usepackage{epsfig}
\usepackage{graphicx}
\usepackage{amsmath}
\usepackage{amssymb}
\usepackage{booktabs}
\usepackage{soul}
\usepackage{dirtytalk}
\usepackage[dvipsnames]{xcolor}
\usepackage[numbers,sort]{natbib} %
\usepackage{balance}
\usepackage{fancyvrb}

\newcommand{\testrec}{Test$^{\text{Rec}}_{2K}$}
\newcommand{\testrecnew}{Test$^{\text{Rec}}_{37K}$}
\newcommand{\testloc}{Test$^{\text{Loc}}_{7K}$}

\def\sepappendix{0}

\usepackage[pagebackref=true,breaklinks=true,colorlinks,bookmarks=false]{hyperref}

\def\cvprPaperID{1204} %
\def\confYear{CVPR 2021}

\begin{document}

\title{Read and Attend: \\ Temporal Localisation in Sign Language Videos}

\author{G\"ul Varol$^{1,2}$\thanks{Equal contribution} \quad Liliane Momeni$^{1*}$ \quad Samuel Albanie$^{1*}$ \quad Triantafyllos Afouras$^{1*}$ \quad Andrew Zisserman$^1$\\
$^{1}$ Visual Geometry Group, University of Oxford, UK \\
$^{2}$ LIGM, \'Ecole des Ponts, Univ Gustave Eiffel, CNRS, France \\
{\tt\small \{gul,liliane,albanie,afourast,az\}@robots.ox.ac.uk}\\
{\tt\small \url{https://www.robots.ox.ac.uk/~vgg/research/bslattend/} }
}

\maketitle

\begin{abstract}

The objective of this work is to annotate sign instances across a broad vocabulary in continuous sign language. We train a Transformer model to ingest a continuous signing stream and output a sequence of written tokens on a large-scale collection of signing footage with weakly-aligned subtitles. We show that through this training it acquires the ability to attend to a large vocabulary of sign instances in the input sequence, enabling their localisation. Our contributions are as follows: (1) we demonstrate the ability to leverage large quantities of continuous signing videos with weakly-aligned subtitles to localise signs in continuous sign language; (2) we employ the learned attention to \textit{automatically} generate hundreds of thousands of annotations for a large sign vocabulary; (3) we collect a set of 37K \textit{manually verified} sign instances across a vocabulary of 950 sign classes to support our study of sign language recognition; (4) by training on the newly annotated data from our method, we outperform the prior state of the art on the BSL-1K sign language recognition  benchmark.

\end{abstract}
\section{Introduction}

\begin{figure}
    \centering
    \includegraphics[width=0.46\textwidth]{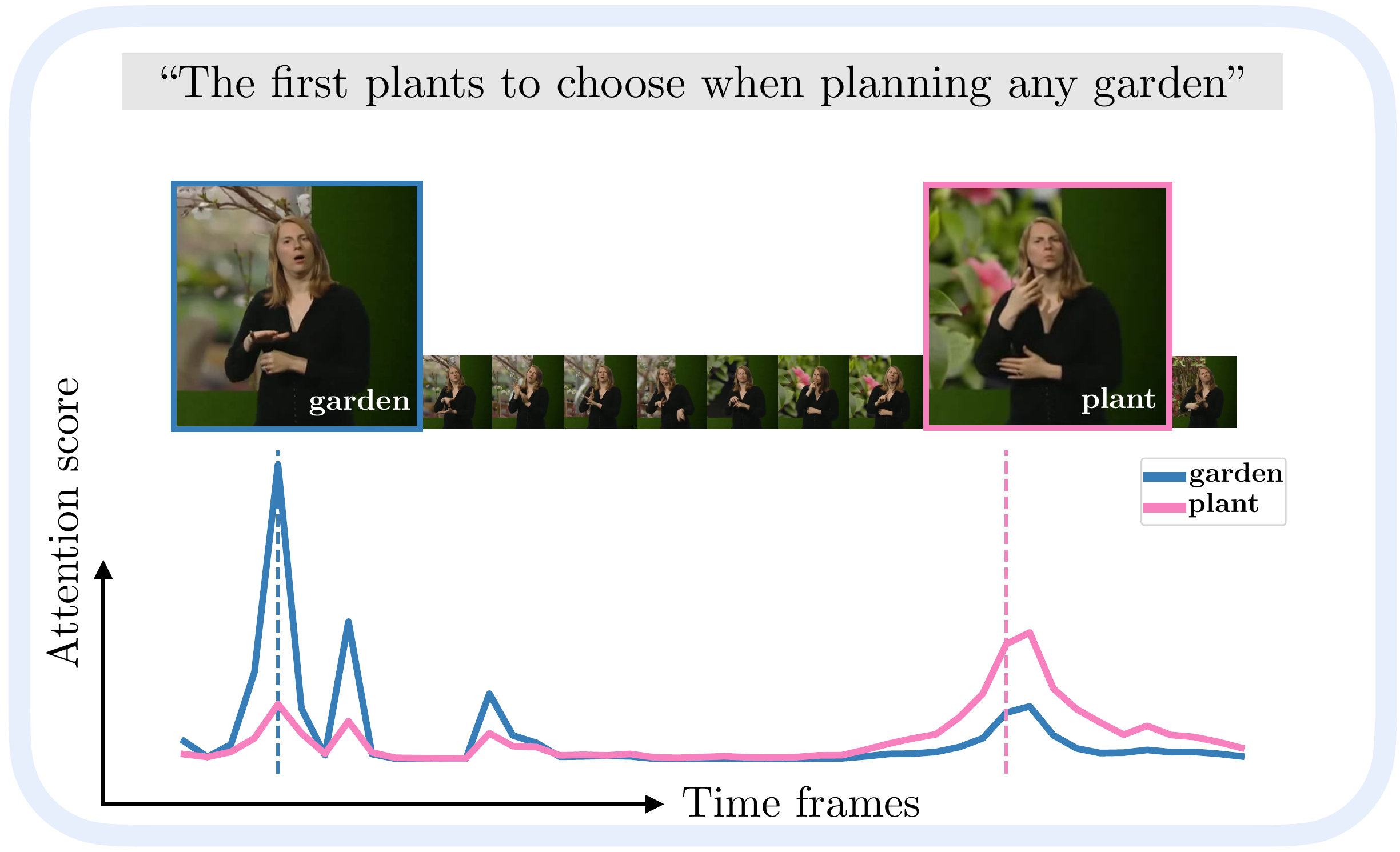}
    \caption{\textbf{Sign localisation emerges from sequence prediction.} In this work, we show that the ability to localise instances of signs emerges naturally by training a Transformer model~\cite{Vaswani2017} to perform a sequence prediction task on hundreds of hours of continuous signing videos with weakly-aligned subtitles. %
    }
    \label{fig:teaser}
\end{figure}

Sign languages are visual languages that, for deaf communities, represent the natural means of communication~\cite{sutton_spence_woll_1999}.
Our goal in this paper is to identify and temporally localise instances of signs among sequences of continuous sign language. Achieving automatic sign localisation enables a diverse range of practical applications: construction of sign language dictionaries to support language learners, indexing of signing content to enable efficient search and \say{intelligent fast-forward} to topics of interest, automatic sign language dataset construction, \say{wake-word} recognition for signers~\cite{rodolitz2019accessibility} and tools to assist linguistic analysis of large-scale signing corpora.

In recent years, there has been a great deal of progress in temporally localising human actions within video streams~\cite{shou2016temporal,zhao2019hacs} and spotting words in spoken languages through aural~\cite{coucke2019efficient} and visual~\cite{Stafylakis17,Momeni20} keyword spotting methods. In both cases, a key driver of progress has been the availability of large-scale annotated datasets, enabling the powerful representation learning abilities of convolutional neural networks to be brought to bear on the task.  

By contrast, annotated datasets for sign language are limited in scale and typically orders of magnitude smaller than their spoken counterparts~\cite{bragg2019}. Widely used datasets such as RWTH-PHOENIX~\cite{koller2015continuous,Camgoz2018} and the CSL dataset~\cite{huang2018video}
provide continuous sign annotations in the form of \textit{glosses}\footnote{Glosses are atomic lexical units used to annotate sign languages.} or free-form sentences, but lack precise temporal annotations and are limited in content diversity, vocabulary,
and scale. %
Large-scale collections of continuous signing videos exist, but %
are limited to sparse annotation coverage~\cite{Albanie20,schembri2013building}.

In the absence of large-scale annotated training data, in this work we turn to a readily available and large-scale source: sign-interpreted TV broadcast footage together with subtitles of the corresponding speech in English. 
We propose to annotate this data with signs by training a Transformer \cite{Vaswani2017} 
to predict, given input streams of continuous signing,  the corresponding subtitles, and then using its trained attention mechanism to perform alignment from English words to signs.

This is a very challenging task: first, subtitles are only \textit{weakly aligned} to the signing content---a sign may appear several seconds before or after its corresponding translated word appears in the subtitles, thus subtitles provide a relatively imprecise cue about the temporal location of a sign. Second, sign interpreters produce a \textit{translation} of the speech that appears in subtitles, rather than a \textit{transcription}---words in the subtitle may not correspond directly to individual signs produced by interpreters, and vice versa. Third, grammatical structures between sign languages and spoken languages differ considerably~\cite{sutton_spence_woll_1999}, and consequently the \textit{ordering} of words in the subtitle is typically not preserved in the signing.

The core hypothesis motivating this approach is that \textit{in order to solve the sequence prediction task, the attention mechanism of the Transformer must be capable of localising sign instances}.
We demonstrate that by employing recent sign spotting techniques~\cite{Albanie20,momeni20watchread} to coarsely align subtitles, sequence prediction is rendered tractable. One of the primary findings of this work is that,
when performed at large scale (across hundreds of hours of continuous signing content),
the ability to localise signs indeed emerges from the attention patterns of the sequence prediction model (Fig.~\ref{fig:teaser}).

We make the following four contributions:
(1) by training on an appropriate sequence prediction task, we show that the attention mechanism of the Transformer learns to attend to specific signs, enabling their \textit{localisation};
(2) we employ the learned attention to \textit{automatically} generate hundreds of thousands of annotations for a large sign vocabulary;
(3) we collect a set of 37K \textit{manually verified} sign instances across a vocabulary of 950 sign classes to support our study of sign language recognition;
(4) by training on the newly annotated data from our method, we outperform the prior state of the art on the BSL-1K sign language recognition  benchmark.

\section{Related Work}

Our approach relates to prior work on
sign language recognition, translation, spotting,
and in particular automatic annotation of sign language data.
We present a discussion of these, followed by a brief
overview of Transformers in natural language processing (NLP)
and works in other domains
using attention mechanisms for localisation.

\noindent \textbf{Sign language recognition and translation.} The 
computer vision community has a long history of efforts to develop 
systems for sign language recognition, reaching back to the 
1980s~\cite{tamura1988recognition}. Initial work focused on 
hand-crafting features~\cite{tamura1988recognition,fillbrandt2003extraction} to model discriminative shape and motion cues and explored 
their usage in combination with Hidden-Markov 
Models~\cite{starner1995visual,vogler2001framework}. These works were
followed by approaches that employed pose estimation as a basis for 
recognition~\cite{ong2012,pfister2014domain}. The community later 
transitioned to employing convolutional neural networks (CNNs) for 
appearance modelling~\cite{Camgoz17}. In particular, the I3D
architecture, originally developed for action 
recognition~\cite{carreira2017quo}, has proven to be effective for 
sign recognition~\cite{Li19wlasl,li2020transferring,Joze19msasl,albanie2020seehear,Momeni20}---we similarly employ this model in 
our work.

\textit{Continuous} sign language recognition entails important challenges compared to \textit{isolated} sign recognition, including epenthesis effects and co-articulation~\cite{bragg2019} as well as the
non-trivial definition of temporal boundaries between signs~\cite{brentari2009effects}.
Towards dealing with these problems, \cite{cheng2020} uses the CTC loss~\cite{Graves2006ConnectionistTC} to infer an alignment between sequence-level annotations and visual input and introduces an 
auxiliary loss to use the alignments as pseudolabels; while
\cite{bull2020} proposes a graph convolutional network to automatically segment large sign language video sequences into short sentences, aligned with their subtitle transcription.

Recent works have applied sequence-to-sequence models to sign language translation.
Camg\"{o}z et al.~\cite{Camgoz2018} use a two-stage pipeline that translates a video into gloss sequences then those into spoken language. Subsequent work~\cite{camgoz2020sign} replaces this framework with a Transformer model trained on frame-level
features jointly for recognition and translation,
while
\cite{camgoz2020multichannel} combines multiple articulators including face and upper body pose to train a translation system without gloss annotations.
These approaches~\cite{Camgoz2018,camgoz2020sign,camgoz2020multichannel}
have shown improvements towards translation in the restricted domain of discourse
of the RWTH-PHOENIX-Weather-2014T German Sign Language (DGS) dataset~\cite{Camgoz2018}.
Ko et al.~\cite{ko2019neural} train a sequence-to-sequence model using keypoint features on Korean Sign Language translation.
Although these methods show promising results in constrained conditions,
open-vocabulary sign language translation in the wild remains largely unsolved. 

\begin{figure*}
    \centering
    \includegraphics[width=.95\textwidth,trim={6cm 4.4cm 9cm 15.7cm},clip]{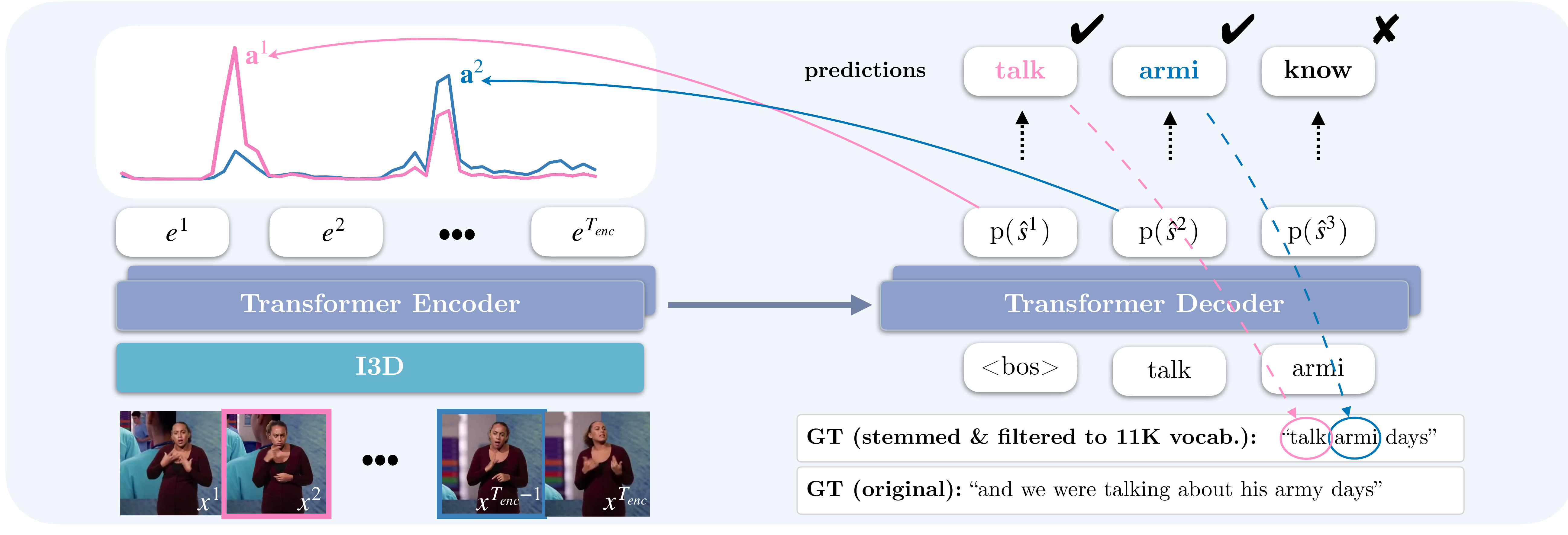}
    \caption{\textbf{Pipeline:} We use an I3D model pretrained on sign classification to extract spatio-temporal visual features by using a sliding window. %
    We then train a 2-layer Transformer model to predict stemmed subtitles from the input video feature sequence. We use the learned model's attention vectors to spot new instances of signs by checking which words in the predicted hypothesis overlap with the stemmed subtitle. For example, here the tokens ``talk" and ``armi", found in the model's hypothesis, also appear in the subtitle and are therefore retained, while ``know" does not and is hence discarded. The location of a new spotting is determined by the index at which the corresponding encoder-decoder attention peaks. Note: we omit the sample index, subscript $i$, shared by all variables (described in Sec.~\ref{sec:method}).
    } 
    \label{fig:pipeline}
    \mbox{}\vspace{-0.8cm}\\
\end{figure*}

\noindent \textbf{Automatic annotation of sign language data.}
Sign language datasets either offer isolated gloss-level annotations of single signs, e.g., MSASL~\cite{Joze19msasl}, WLASL~\cite{Li19wlasl}, or are heavily constrained in visual domain and vocabulary, e.g., RWTH-PHOENIX~\cite{koller2015continuous,Camgoz2018}, KETI~\cite{ko2019neural} (only 105 sentences). 
Large-scale continuous sign language datasets, on the other hand,
are not exhaustively annotated \cite{Albanie20,bslcorpus17}.
The recent efforts of Albanie et al.~\cite{Albanie20}
scale up the automatic annotation of sign language data,
and construct the BSL-1K dataset with the help of a visual keyword spotter~\cite{stafylakis2018zero,Momeni20} trained on lip reading to detect
instances of mouthed words as a proxy for spotting signs.
\textit{Sign spotting} refers to a specialised form of sign language recognition in which the objective is to find whether and where a given sign has occurred within a sequence of signing.
It has emerged as an intermediate step to collect more annotated sign language data. 
With this goal, Momeni et al.~\cite{momeni20watchread} use
dictionary lookups in subtitled videos and improve low-shot sign spotting.
Other automatic annotation approaches include an automatic pipeline for active signer
detection and sign language diarisation~\cite{albanie2020seehear}. While these previous methods are\textit{ context-free}, in this work, we introduce a \textit{context-aware} approach that can be used to localise signs automatically. In fact, while we profit from
annotations obtained in prior works using mouthing cues~\cite{Albanie20} and dictionaries~\cite{momeni20watchread}, our approach differs
considerably from theirs in method---we define the supervision directly on subtitles
and formulate the problem as a sequence-to-sequence prediction task. We demonstrate the benefits of our approach empirically in Sec.~\ref{sec:experiments}.

\noindent\textbf{Transformers in NLP.}
Incorporating an attention mechanism into encoder-decoder architectures led to a revolution in neural machine translation \cite{bahdanau2014neural} by reducing dependency on strong text alignment.
Vaswani et al.~\cite{Vaswani2017} further extended this approach by replacing all recurrent and convolutional components of a sequence-to-sequence model with self-attention. 
Even though such methods implicitly model source-to-target alignment with attention, their primary focus is on translation performance, rather than word-alignment.
\cite{Garg2019} further studies how to simultaneously optimise for accurate word-alignment without sacrificing translation performance---we investigate a variant of their approach in Sec.~\ref{sec:experiments}. 

\noindent \textbf{Attention mechanisms for localisation.}
Cross-modal attention has been employed in the literature for various localisation problems such as visual grounding in videos~\cite{xu2019multilevel,yuan2019find,liu2018attentive,chen2018temporally} or images~\cite{deng2018visual,yu2018mattnet}, keyword spotting in audio~\cite{shan2018attention} or visual speech~\cite{stafylakis2018zero,Momeni20} and audio-visual sound source localisation~\cite{Arandjelovic18objects, senocak2018learning, harwath2018jointly}. However, to the best of our knowledge, our work is the first to apply these ideas at large-scale to sign localisation from weakly-aligned subtitles.

\section{Sign Localisation with Attention}
\label{sec:method}

In this section, we describe how we train a Transformer model
on a weakly-supervised sign language sequence-to-sequence task 
and then use the trained model to perform sign localisation (see Fig.~\ref{fig:pipeline}
for an overview).

Let $\mathcal{X}_{\mathfrak{L}}$ denote the space of sign language video segments $\mathfrak{L}$,  and $\mathcal{T}$ denote the space of subtitle sentences. Further, let  $\mathcal{V}_{\mathfrak{L}}=~\{1, \dots, V \}$ represent the \textit{vocabulary} (an enumeration of spoken language tokens that correspond to signs that can be performed in $\mathfrak{L}$) and let $\mathcal{S}$ denote a subtitled collection of $I$ videos containing continuous signing, $\mathcal{S} = \{(x_i, s_i ) : i \in \{1, \dots, I\}, x_i \in \mathcal{X}_{\mathfrak{L}}, s_i \in \mathcal{T}\}$. Our objective is to localise potential occurrences of signs in $\mathcal{S}$.

\vspace{0.2cm}
\noindent\textbf{Transformer training with subtitled videos.}
To address this task, we propose to train a sequence-to-sequence model with attention.
Given a video-subtitle pair $(x_i,s_i) \in \mathcal{S}$, we train 
a Transformer~\cite{Vaswani2017} to predict
the target text sequence $s_i = (s_i^1, s_i^2 \cdots, s_i^{T_{dec}})$ from the source video sequence  $x_i = (x_i^1, x_i^2, \cdots x_i^{T_{enc}})$, one token at a time.
Specifically, the Transformer's encoder transforms $x_i$ into an encoded sequence
$enc(x_i) = (e_i^1, e_i^2, \cdots e_i^{T_{enc}})$.
The decoder then attends on the encoded sequence and predicts the output sequence $\hat{s}_i = (\hat{s}_i^1, \hat{s}_i^2, \cdots \hat{s}_i^{T_{dec}})$  auto-regressively,
factorising its joint probability into a product of individual conditionals:
\begin{align}
  p(\hat{s}_i|x_i) =  \prod_{t=1}^{T_{dec}} p(\hat{s}_i^t  | {\hat{s}_i^1,\hat{s}_i^2 \cdots \hat{s}_i^{t-1}}, enc(x_i)).
\end{align}

Using the target subtitles $s_i$ as the ground truth output sequences, we train the model to maximise their log likelihoods by minimising the following loss:
\begin{align}
  \mathcal{L} = - \mathbb{E}_{(x_i, s_i) \in \mathcal{S}}  \log p(s_i|x_i)
  \label{eqn:mil-nce}
\end{align}
Note that we assume access to a sparse collection of automatic sign annotations,  $\mathcal{N} = \{(x_k, v_k) : k \in \{1, \dots, K\}, v_k \in
\mathcal{V}_\mathfrak{L}, x_k \in \mathcal{X}_\mathfrak{L}, \exists (x_i, s_i) \in
\mathcal{S} \, s.t. \, x_k \subseteq x_i \}$, using mouthing cues~\cite{Albanie20} and dictionaries~\cite{momeni20watchread}. In practice, we restrict the Transformer training on a subset of videos $ \mathcal{S}_A \subseteq \mathcal{S}$, containing at least one of these annotations within the subtitle timestamps,
formally 
$\mathcal{S_A} = \{ (x_a,s_a) : a \in \{1, \dots, A\}, x_a \in \mathcal{X}_\mathfrak{L}, \exists (x_k, s_k) \in
  \mathcal{N} \, s.t. \, x_k \subseteq x_a\}$. This ensures approximate alignment between the source video and target subtitle. For arbitrary sequences in $\mathcal{S}$ this is not guaranteed due to imperfect synchronisation between subtitles (corresponding to audio) and sign language interpretation. The goal of our training is therefore to exploit the knowledge of the unannotated words in the subtitles in $\mathcal{S}_A$ in order to discover a new collection of  $(x,v)$  sign-video pairs  (that is not included in $\mathcal{N}$)  in the entire set $\mathcal{S}$. %

\vspace{0.2cm}
\noindent\textbf{Localising new sign instances with attention.}
Next, we describe how we use the Transformer model
to look for new sign instances (see Fig.~\ref{fig:pipeline}).
After inputting the video sequence $x_i$ into the trained model, we use a decoding strategy (e.g., greedy) to predict the output sequence  $\hat{s}_i$  and corresponding attention vectors $ a_i = (\textbf{a}_i^1, \textbf{a}_i^2, \cdots
\textbf{a}_i^{T_{dec}}) \in R^{T_{dec} \times T_{enc}}$.
We iterate over the predicted sequence $\hat{s}_i$ and localise new sign instances \textit{only} for the tokens
predicted correctly (i.e., appearing in subtitle $s_i$);
the video location is determined by the index at which the
corresponding attention vector is maximised, to yield sets of (location, sign) pairs of the form:
$\{ (\text{argmax}_{j\in\{1,2 \cdots T_{enc}\}} \textbf{a}_i^t(j), s_i^t) : \hat{s}_i^t = s_i^t,  t \in \{1,2 \cdots T_{dec}\}  \}$.

\vspace{0.2cm}
\noindent\textbf{Implementation details.}
We represent the input video $x_i$
with features extracted using a pretrained
spatio-temporal
convolutional neural network model, applied in a sliding window manner with a 4-frame stride. In particular,
we train an I3D architecture~\cite{carreira2017quo}
on an extended set of automatic annotations  $\mathcal{N}$
that we obtain by combining the methods of \cite{Albanie20}
and \cite{momeni20watchread}, to spot signs via
mouthing cues and sign language dictionaries, respectively.
We train with a single-sign classification objective and
follow the same hyperparameters (e.g., 16-frame
inputs) of the sign language
recognition models in~\cite{Albanie20}. The 1024-dimensional video features from I3D are used as input to the Transformer encoder.

To construct ground-truth text labels for
our Transformer training, we stem the words in every subtitle
under the assumption that variations of a written word
could map to the same sign. We note that the many-to-many
mapping between words and signs is a complex problem,
which we do not explicitly deal with in this work.
To establish a tractable problem,
we define a vocabulary of 11,515 stems based on their
frequency and occurrence within the automatic annotations  $\mathcal{N}$.
This is reduced from an original set of 40K words appearing
in the full set of subtitles $S$. We further remove
stop words for which there is often no sign correspondence.
This approach resembles \textit{glossing}
sign language data, i.e., representing sign
sequences with word sequences, without spoken language
grammar.

Following common practice in the sequence-to-sequence literature~\cite{Vaswani2017}, we train the model with teacher forcing~\cite{williams1989learning}, i.e.\ at every decoding step we provide the  previous-step's ground truth as input to the decoder.
During inference we experiment with three different decoding strategies: auto-regressive greedy decoding, left-to-right beam search, and teacher forcing.
With greedy decoding, we iterate over the available sequences and for each one,
we select as new spottings
all the words in the predicted hypothesis that appear in the reference subtitle.
For beam search, we iterate over the predictions which overlap with the reference from the multiple returned hypotheses, and select for each predicted word the location with maximum attention score.
We show results for another variant of beam search where we choose the hypothesis with the highest recall in the appendix
\if\sepappendix1{(Sec.~C.3).}
\else{(Sec.~\ref{subsec:app:decoding}).}
\fi
With teacher forcing, we do not use the token predictions of the model, but only the attention scores, which we associate with the next ground-truth word in the subtitle at every decoding step.
Since we consider all words in the subtitles, this strategy provides good yield but no notion of the model's confidence. In order to obtain a confidence score we use the following heuristic:
For every sequence, a word found in the subtitle is automatically annotated if the attention peak for the corresponding decoding step is higher than a threshold $\tau$.

When using Transformers with multiple attention heads, we obtain single attention scores by averaging the attention vectors of the individual heads. In Sec.~\ref{sec:dec_layer_attentions} we discuss results on combining attention from different decoder layers.

\section{Experiments}
\label{sec:experiments}

This section is structured as follows:
We first present the datasets used as well as the various training and evaluation protocols
that we follow in our experiments (Sec.~\ref{subsec:datasets}).
Next, we show how we choose our pretrained input video features (Sec.~\ref{subsec:features}).
Then, we evaluate our Transformer models trained with these features
and 
discuss different strategies for mining new instances to obtain an automatically
annotated training set (Sec.~\ref{subsec:mining}).
We show that, when adding our newly mined training samples,
we outperform the previous state of the art on sign language recognition (Sec.~\ref{subsec:downstream}).
Finally, we provide qualitative results on two datasets (Sec.~\ref{subsec:qualitative})
and discuss limitations (Sec.~\ref{subsec:discussion}).

\subsection{Data and evaluation protocols}
\label{subsec:datasets}

\noindent\textbf{Datasets.} We use BSL-1K~\cite{Albanie20}, a large-scale, subtitled and sparsely annotated dataset (for a vocabulary of 1,064 signs) of more than 1000 hours of continuous signing from sign language interpreted BBC television broadcasts. The programs cover a wide range of genres: from medical dramas and nature documentaries to cooking shows. In Sec.~\ref{subsec:qualitative}, we show qualitative examples on the RWTH-PHOENIX~\cite{Camgoz2018} dataset, which is significantly smaller in size and from weather broadcasts only, restricting the domain of discourse.

\begin{table}
    \setlength{\tabcolsep}{1pt}
    \centering
    \resizebox{0.999\linewidth}{!}{
        \begin{tabular}{lr|cccc|cccc}
            \toprule
            & & \multicolumn{4}{c}{\testrec{}~\cite{Albanie20}} & \multicolumn{4}{|c}{\testrecnew{}}\\
            \midrule
            & & \multicolumn{4}{c}{2K inst. / 334 cls.} & \multicolumn{4}{|c}{37K inst. / 950 cls.}\\
            \midrule
            & & \multicolumn{2}{c}{per-instance} & \multicolumn{2}{c}{per-class} & \multicolumn{2}{|c}{per-instance} & \multicolumn{2}{c}{per-class} \\
            Training & \#ann. & top-1 & top-5 & top-1 & top-5 & top-1 & top-5 & top-1 & top-5 \\
            \midrule
            M~\cite{Albanie20}$\mathsection$ & 169K  & 76.6 & 89.2 & 54.6 & 71.8 & 26.4 & 41.3	& 19.4 & 33.2 \\
            D & 510K & 70.8 & 84.9 & 52.7 & 68.1 & 60.9 & 80.3 & 34.7 & 53.5 \\
            M+D & 678K & \textbf{80.8} & \textbf{92.1} & \textbf{60.5} & \textbf{79.9} & \textbf{62.3} & \textbf{81.3} & \textbf{40.2} & \textbf{60.1} \\
            \bottomrule
        \end{tabular}
    }
    \vspace{2pt}
    \caption{\textbf{A new recognition test set \testrecnew{} and an improved I3D model:}
    We employ the method of~\cite{momeni20watchread} to find
    signs via automatic dictionary spotting (D), significantly
    expanding the training and testing data obtained from mouthing cues by~\cite{Albanie20} (M).
    We also significantly expand the test set by manually
    verifying these new automatic annotations from the test partition
    (\testrec{} vs \testrecnew{}).
    By training on the extended M+D data, we obtain state-of-the-art results,
    outperforming the previous work of~\cite{Albanie20} and providing strong I3D features for the subsequent steps of our method.
    $\mathsection$The slight improvement in the performance of \cite{Albanie20}
    over the original results reported in that work is due to 
    our denser test-time averaging when
    applying sliding windows (8-frame vs 1-frame stride).
    }
    \label{tab:i3d}
\end{table}

\noindent\textbf{Transformer training and evaluation on \testloc{}.} To form the video-subtitle
training data pairs, we sample 183K ($\mathcal{S}_A$) out of 685K subtitles from the
BSL-1K training set ($\mathcal{S}$), in which there exists 
at least 1 automatic annotation (with a confidence score above 0.7) from the annotations collection $\mathcal{N}$. $\mathcal{N}$ is formed by applying the method of~\cite{Albanie20} on a large vocabulary of words beyond 1K to find signs via mouthing cues and applying the method of~\cite{momeni20watchread} to find signs via automatic dictionary spotting.
See appendix
\if\sepappendix1{(Sec.~C.2)}
\else{(Sec.~\ref{subsec:app:md})}
\fi
for details on this step.
Subtitles originally contain 9.8 words from the initial 40K words vocabulary on average, which is reduced to 4.4 words per subtitle from the 11K stems vocabulary after stemming and filtering. Corresponding videos are tightly extracted according to the subtitle
timestamps, and are on average 3.52 seconds long.

For evaluating the localisation capability of the proposed method,
we use the automatic annotations $\mathcal{N}$ in the BSL-1K test set whose confidence scores
are above 0.9, resulting in 7497 subtitle-video pairs with a total of
7661 annotations, referred to as \testloc{}.
We measure the localisation accuracy for the annotated words in each subtitle and only on the correct predictions: we consider a correct prediction to be also correctly localised if its predicted location lies within 8 frames of the annotation time.
We also report recall and precision of the model's predictions for each sequence
by measuring
the percentage of words in the subtitle that are predicted (recall)
and the percentage of predicted words which appear in the subtitle (precision).
For all three metrics, we report the average over all sequences in the test set.

\noindent\textbf{Single-sign recognition benchmark.} %
In order to justify the value of our
automatic annotation approach with the Transformer model,
we evaluate on the proxy task of single-sign
recognition on trimmed videos by using our localised sign instances 
from the training set as labels for classification training.
Similar to~\cite{Albanie20,Joze19msasl,Li19wlasl}, we adopt
top-1 and top-5 accuracy metrics reported with
and without class-balancing.

We use the BSL-1K manually verified recognition test set with 2K samples~\cite{Albanie20},
which we denote with \testrec{},
and significantly extend it to 37K samples as \testrecnew{}.
We do this by collecting new annotations from human annotators using the VIA 
tool~\cite{dutta2019via}
with a verification task as in~\cite{Albanie20}. This extended test set %
reduces the bias towards signs with
easily spotted mouthing cues
(since we also include dictionary spottings~\cite{Momeni20})
and spans a larger fraction of the training
vocabulary, i.e.~950 out of 1064 sign classes
(vs 334 classes in the original benchmark \testrec{} of~\cite{Albanie20}).

\subsection{Comparison of video features}
\label{subsec:features}
We first conduct experiments to determine which I3D video features are best suited 
as input to the Transformer model
as described in Sec.~\ref{sec:method}.
In Tab.~\ref{tab:i3d}, we demonstrate the benefits of combining
annotations from both mouthing (M)~\cite{Albanie20}
and dictionary spottings (D)~\cite{momeni20watchread}.
We show that our sign
classification training using 678K automatic
annotations obtains state-of-the-art performance
on \testrec{},
as well as our new and more challenging test set \testrecnew{}.
We therefore use this M+D model for the rest of our experiments.
Note that all three models in Tab.~\ref{tab:i3d} (M, D, M+D) are pretrained on
Kinetics~\cite{carreira2017quo},
followed by video pose distillation as described in~\cite{Albanie20}.
We observed no improvements when initialising
M+D training from M-only pretraining.

\begin{table}
    \setlength{\tabcolsep}{1pt}
    \centering
    \resizebox{0.99\linewidth}{!}{
        \begin{tabular}{lcc|rc|rc}
            \toprule
            & & & \multicolumn{2}{c|}{Loc. Acc. (GD)} & \multicolumn{2}{c}{Loc. Acc. (TF)} \\
            Tr. & Recall & Prec. & Att. layer 1/2/3 & [avg] & Att. layer 1/2/3 & [avg] \\
            \midrule
            1L & 15.8 & 36.4          & 65.9             & [65.9]          & 44.8 & [44.8] \\
            2L & \textbf{16.5} & \textbf{37.2} & 63.9/57.8        & [\textbf{66.1}] & 51.1/37.6 & [44.5] \\
            3L & \textbf{16.5} & 36.9 &  62.5/60.8/16.4 & [65.3]   & \textbf{51.4}/38.4/15.7 & [46.4]\\
            \bottomrule
        \end{tabular}
    }
    \vspace{3pt}
    \caption{
    \textbf{Localisation performance of attention layers.}
    We evaluate the performance of Transformers on \testloc{}
    for different number of encoder/decoder layers in the training (different rows).
    We report the localisation accuracy for the encoder-decoder attention
    scores from every layer, as well as the average over layers,
    for both teacher forcing (TF) and greedy decoding (GD) modes.
    }
    \label{tab:subtitle_training}
\end{table}

\subsection{Mining training examples through attention} \label{subsec:mining}

Next, we ablate different design choices for the Transformer model.

\noindent\textbf{Which attention layer for sign-video alignment?} \label{sec:dec_layer_attentions} 
Similarly to~\cite{Garg2019}, we conduct an investigation into
which decoder layer gives attention scores that are more useful for localising signs. 
We train three models, with 1, 2 and 3 encoder and decoder layers and report the localisation accuracy when using the attention from each layer separately, or an average of all layers. 
The results on \testloc{} %
in Tab.~\ref{tab:subtitle_training} suggest that averaging the attention scores 
over all layers gives the best localisation when using greedy auto-regressive decoding, while 
using the attention scores from the first decoder layer works best with teacher forcing. 
We note that this finding stands in contrast to those of~\cite{Garg2019} which 
concluded that the penultimate layer works better for word alignment in a machine 
translation task.
We conjecture that the difference results from the different nature of
the two domains, i.e., video versus text inputs.
In terms of precision and recall, all three models perform
similarly with rates at 37\% and 16\%, respectively.
We continue with a 2-layer Transformer model for the rest of the experiments
and
given the observations in Tab.~\ref{tab:subtitle_training},
we use the layer-averaged attention with greedy decoding
and the first layer attention with teacher forcing.

\noindent\textbf{Incorporating sparse annotations.}
As explained in Sec.~\ref{sec:method},
we make use of the available sparse annotations $\mathcal{N}$
to restrict the training subtitles to those with at least 1 annotation.
When removing this constraint, the model does not train as well,
and reaches a recall of only 
6.8\% (vs 16.5\%).

Here, we also report some of our findings by employing three
additional strategies
to improve the Transformer training using the
sparse annotations $\mathcal{N}$. In all three cases, we observe
no or minor gains (on \testloc{}), at the cost of a more complex
method and the need for annotations. Therefore, we do not integrate
them in our final model
and provide detailed results in appendix
\if\sepappendix1{(Sec.~C.2).}
\else{(Sec.~\ref{subsec:app:sparse}).}
\fi

\textit{Alignment loss on sparse annotations:}
We investigate whether the sparse annotations $\mathcal{N}$ could be used for supervising the sign-video alignment
explicitly (similar to~\cite{Garg2019} in NLP).
To this end, we define an additional
loss that operates on the encoder-decoder
attention to enforce a high response
whenever there is known location information.
We achieve this via an additional L2 loss term
between a 1D gaussian centered around the 
annotated time frame and the corresponding
attention vector. While the localisation
performance with teacher-forcing
increases (58.7\% vs 51.1\%), it still remains lower compared to the corresponding greedy decoding result and
we observe no significant gains
for other metrics measured on the predictions.

\textit{Curriculum learning with sparse annotations:}
To provide warmup for the model training,
we start by temporally trimmed video inputs
around known sign locations $\mathcal{N}$. We gradually
increase the number of annotations from 1 to 3,
before we fully input the subtitle duration
to the Transformer.
We only observe minor improvements:
16.0\% vs 15.8\% recall with the 1-layer architecture.

\textit{Subtitle alignment through active signer
detection and sparse annotations:}
To overcome the alignment noise present in the data,
we apply an algorithm that combines a pose-based
active signer detection~\cite{albanie2020seehear}
and the knowledge of sparse annotations $\mathcal{N}$.
Specifically, we apply temporal shifts to subtitles
such that their temporal midpoint aligns with the
average time of any annotated signs they contain.
We then apply affine transformations to the subtitles
without annotations such
that they fill the regions between those with annotations,
subject to the
hard constraint that the expansions do not overlap periods of inactive signing.
This approach increases the amount of training subtitles
with annotations to 230K; however, training with this
new set does not improve recall (15.4\% vs 16.5\% with 2-layers).

\begin{table}
    \setlength{\tabcolsep}{2pt}
    \centering
    \resizebox{0.99\linewidth}{!}{
        \begin{tabular}{lrrr|rr}
            \toprule
                             & \#subtitles  & \#ann. & \#ann. & top-1   & top-1 \\
            Spotting mode    & unannot.   &  11K &   1K &         per-inst   & per-cls\\
            \midrule
            TF ($\geq .2$)    & 114K & 290K &  97K & 22.2  &   4.7      \\
            TF ($\geq .1$)    & 408K & 1.7M & 545K & 37.3  &  13.4      \\
            TF ($\geq .05$)   & 457K & 2.3M & 754K &  38.7  & 14.4  \\
            TF ($\geq .05$) (align. loss) &457K & 2.3M  & 757K & 38.8 & 14.6 \\
            \midrule
            BS (10 best)        & 109K  & 329K & 166K & 49.6  & 22.7\\
            GD (no subtitle filtering)   & 480K & 1.4M & 910K & 50.6  & 22.6  \\
            GD (align. loss)    & 53K  & 188K & 108K & 53.6 & \bf{24.8} \\
            GD                  & 53K  & 188K & 107K & \bf{53.9}  & \underline{24.7} \\
            \bottomrule
        \end{tabular}
    }
    \vspace{3pt}
    \caption{
    \textbf{Automatically annotating the training data:}
    We show the yield obtained from various decoding strategies
    in terms of number of additional annotations (left).
    Training models only with these annotations,
    we evaluate the
    recognition accuracy on \testrecnew{}. Greedy decoding (GD) obtains better results than teacher forcing (TF)
    even when not filtering the predictions
    against the ground-truth subtitles. Neither including
    10 best predictions from beam search (BS) nor using
    the model trained with the alignment loss
    influences the recognition evaluation significantly.
    }
    \label{tab:yield}
\end{table}

\noindent\textbf{Which decoding mechanism?}
To form a new annotated set for sign recognition training,
we apply the trained Transformer models on the whole 685K
training video-subtitle pairs of the BSL-1K dataset.
In Tab.~\ref{tab:yield} we summarise and compare the yield of new training samples mined with the different decoding strategies we discussed in Sec.~\ref{sec:method}.
We report the number of previously unannotated subtitles, for
which the attention mechanism is able to localise signs, to
demonstrate the benefits of our approach. We also report
the amount of new annotations for both the full 11K vocabulary
and the 1064-subset which is used for the proxy recognition evaluation.
We observe that a significant number of new automatic
sign annotations are obtained with our approach.

To compare the different decoding strategies, we train recognition models on the resulting training sets containing the new annotations and evaluate them on the proxy sign recognition task.
Note that for faster training, we learn a 4-layer MLP
architecture on top of the pre-extracted I3D video features (architecture and optimisation details are given in the appendix, see
\if\sepappendix1{Sec.~D).}
\else{Sec.~\ref{sec:app:details}).}
\fi

We observe that greedy decoding with the simple
filtering mechanism (checking against ground truth) gives best downstream recognition performance
on \testrecnew{}.
Teacher forcing,
beam search and no filtering all yield larger
but noisier training sets that result in lower performance.
However, we note that the ``no subtitle filtering'' experiment
assumes no access to ground-truth subtitles during annotation mining 
and uses
all the predictions, while providing competitive recognition
performance (50.6\% vs 53.9\%).

\begin{table}
    \setlength{\tabcolsep}{8pt}
    \centering
    \resizebox{0.99\linewidth}{!}{
        \begin{tabular}{lr|cccc}
            \toprule
            & & \multicolumn{2}{c}{per-instance} & \multicolumn{2}{c}{per-class} \\
            Training & \#ann. & top-1 & top-5 & top-1 & top-5  \\
            \midrule
            A & 107K & 54.0$^{\pm0.08}$ & 67.9$^{\pm0.10}$ & 24.8$^{\pm0.10}$ & 35.5$^{\pm0.20}$ \\
            \midrule
            M~\cite{Albanie20}$\dagger$ & 169K & 40.8$^{\pm0.17}$ & 62.2$^{\pm0.07}$	& 21.7$^{\pm0.19}$ & 38.5$^{\pm0.29}$ \\
            M+A & 276K & 58.5$^{\pm0.17}$	& 75.5$^{\pm0.02}$ & 30.4$^{\pm0.04}$ & 45.9$^{\pm0.26}$ \\
            \midrule
            D~\cite{momeni20watchread}$\dagger$ & 510K & 62.1$^{\pm0.24}$ & 80.8$^{\pm0.10}$ & 35.1$^{\pm0.38}$ & 54.3$^{\pm0.11}$ \\
            D+A & 276K & 64.2$^{\pm0.08}$ & 81.7$^{\pm0.07}$ & 36.0$^{\pm0.26}$ & 54.0$^{\pm0.32}$ \\
            \midrule
            M+D & 678K & 63.5$^{\pm0.28}$ & 82.1$^{\pm0.04}$ & 37.2$^{\pm0.12}$ & \textbf{56.4$^{\pm0.17}$} \\
            M+D+A & 786K & \textbf{65.0$^{\pm0.14}$} & \textbf{82.6$^{\pm0.02}$} & \textbf{37.9$^{\pm0.07}$} & 56.3$^{\pm0.02}$ \\
            \bottomrule
        \end{tabular}
    }
    \vspace{5pt}
    \caption{
    \textbf{Sign recognition on BSL-1K \testrecnew{}:} We evaluate our 4-layer
    MLP classification models trained on video feature inputs for 1064-sign recognition for various training label sets:
    mouthing~(M), dictionary~(D), and our proposed attention~(A) spottings.
    We obtain state-of-the-art results,
    by consistently improving
    over previous works when including our attention localisations.
    $\dagger$The results are obtained
    from our MLP trained with the annotations from \cite{Albanie20}
    and our application of \cite{momeni20watchread}.
    }
    \vspace{-0.5cm}
    \label{tab:sota}
\end{table}

\subsection{Comparison with other automatic annotations}
\label{subsec:downstream}
\vspace{-0.1cm}
In this section, we train for sign recognition on BSL-1K~\cite{Albanie20} on various label
sets, comparing different automatic annotation methods
and showing that our new sign instances are complementary
when added to training data, achieving
state of the art. As in the previous experiments, we use the MLP
architecture on frozen I3D features to compare the different annotation sets.
This time we perform 3 trainings per model with
different random seeds and report the average and
standard deviation. \\
\indent Tab.~\ref{tab:sota} summarises the results on \testrecnew{}.
We first note that the MLP performance of M+D
annotations matches and slightly outperforms that of I3D from Tab.~\ref{tab:i3d}
(63.5\% vs 62.3\%), validating the suitability of MLP for efficiently comparing annotation set quality.
When compared to the visual keyword spotting
through mouthing (M)~\cite{Albanie20},
our automatic attention localisations (A)
show significant improvements. Furthermore, we observe consistent improvements when combining our new annotations with either the mouthing (M+A) or dictionary (D+A) annotations.
Combining all available annotations (M+D+A),
we achieve state-of-the-art performance (65\%)
outperforming previous work of~\cite{Albanie20} (M: 40.8\%),
as well as a new much stronger baseline (D: 62.1\%)
that we establish in this work,
which uses the new annotations obtained
using sign language dictionaries for sign spotting
\cite{momeni20watchread}.
Our final recognition model can be interpreted as distilling
information from multiple sources (mouthing, dictionary, attention),
each of which has access to a large training set. \\
\begin{figure}
    \centering
    \includegraphics[width=0.47\textwidth]{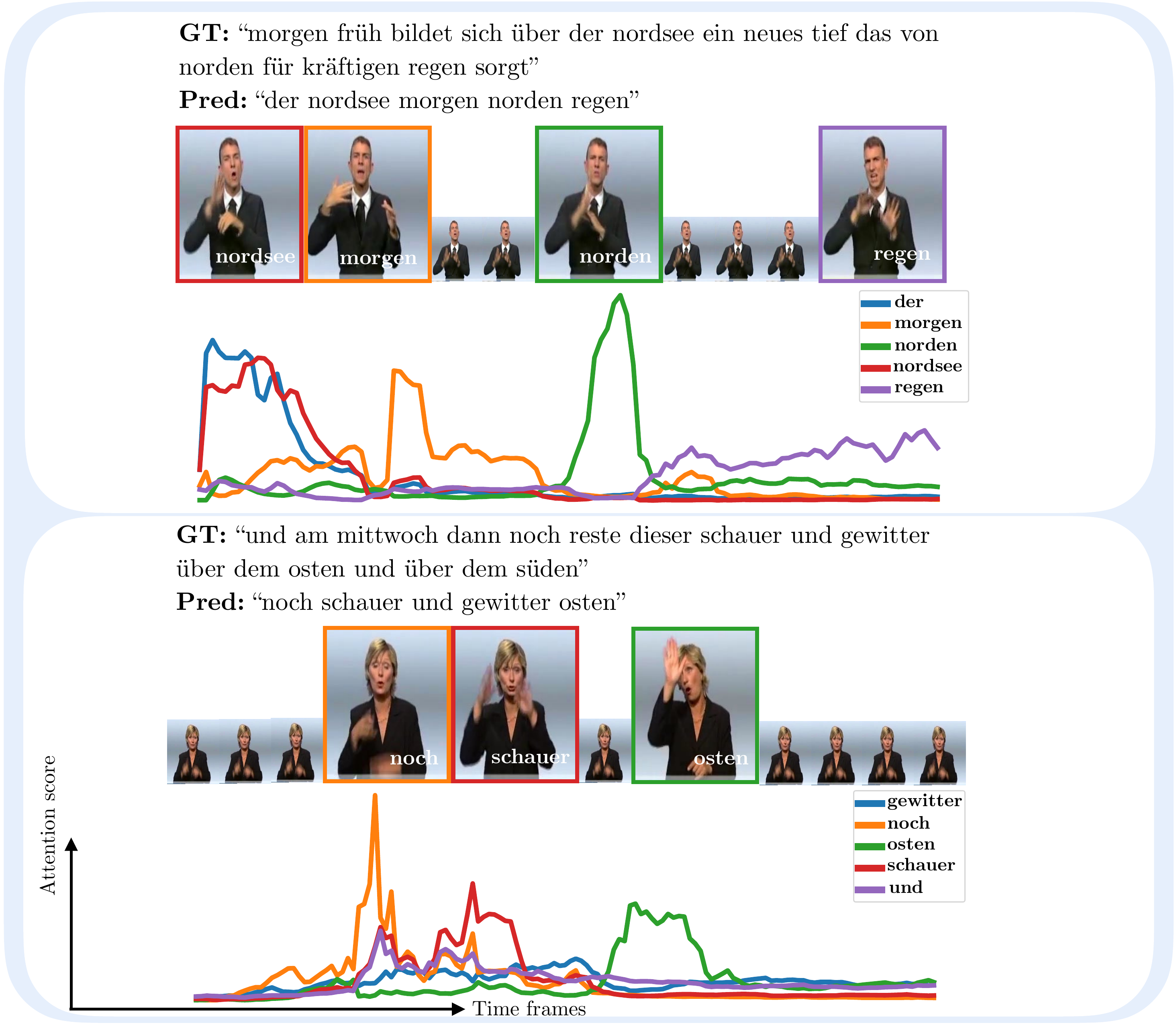}
    \caption{\textbf{Qualitative analysis on the RWTH-PHOENIX:} We show example sign localisation
    results on the test set of RWTH-PHOENIX 2014T. For each video clip, we show the ground-truth sentence as well as the predicted words from the Transformer model of \cite{camgoz2020sign} which overlap with the target sentence. We plot attention scores over time frames for these predicted words and show the frame index at which the corresponding attention vector is maximised for a subset of the correctly predicted words. 
    }
    \label{fig:phoenix}
\end{figure}
\begin{figure*}
    \centering
    \includegraphics[width=0.97\textwidth]{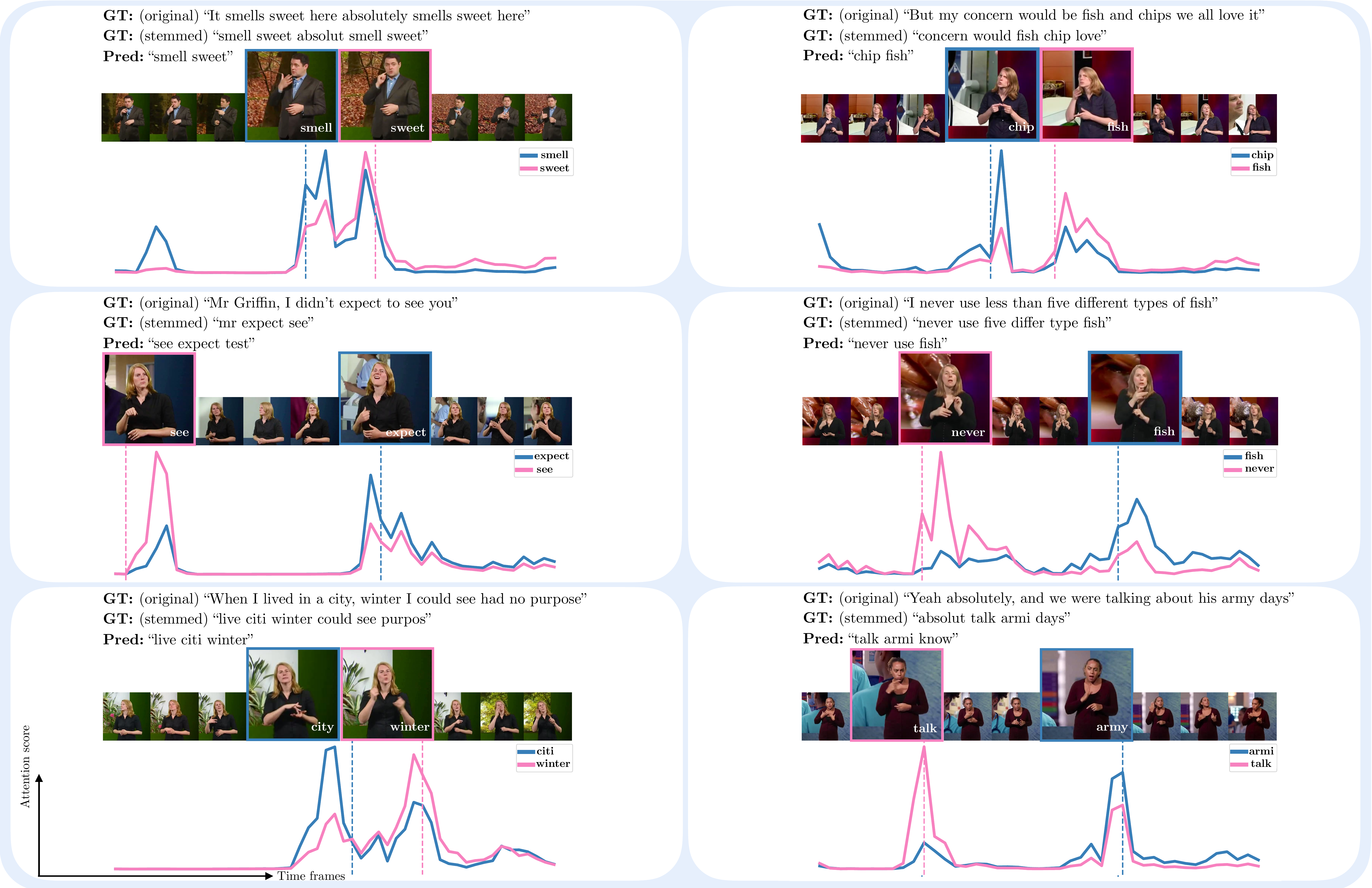}
    \caption{\textbf{Qualitative analysis on BSL-1K:} We show example sign localisation
    results on the BSL-1K test set (\testloc{}). For each video clip, we show the original subtitle,
    the ground-truth stemmed and filtered to 11K vocabulary version, and the prediction of our Transformer model. We plot attention scores over time frames for the predicted words which overlap with the subtitle and for which we have annotated sign times in $\mathcal{N}$ (shown by vertical dashed lines).
    We highlight the frame at which the corresponding attention vector is maximised.
    }
    \label{fig:qualitative}
    \vspace{-0.25cm}
\end{figure*}
\indent We also evaluate the performance of our MLP trained on M+D+A annotations on the BSL-1K sign spotting benchmark proposed by~\cite{Albanie20}, following their protocol, and achieve a score of 0.174 mAP, outperforming the previous state-of-the-art performance of 0.170 mAP \cite{momeni20watchread} and 0.159 mAP~\cite{Albanie20}.

\subsection{Qualitative analysis} %
\label{subsec:qualitative}
 We demonstrate the potential of our Transformer model to localise sign instances through its attention mechanism. Fig.~\ref{fig:qualitative} shows qualitative examples of localising multiple signs, by plotting attention scores over video time frames for predicted words that occur in corresponding subtitles of the BSL-1K test set (\testloc{}). We observe close alignment with the automatic annotations $\mathcal{N}$. One potential limitation of this approach
 for localisation is that the attention vector does not peak only at the corresponding sign location, but also on other signs suggesting that the predictions use context (e.g., ``smell’' and ``sweet’' in Fig.~\ref{fig:qualitative}, top-left).
 
We also investigate whether this localisation ability extends to other datasets. In particular, we reproduce the translation method of Camg\"{o}z et al.~\cite{camgoz2020sign} on RWTH-PHOENIX 2014T~\cite{Camgoz2018} and similarly to \cite{Camgoz2018}, we visualise the attention score plots for predicted words in Fig.~\ref{fig:phoenix}. We are unable to compute the localisation accuracy as sign annotation times are not available for RWTH-PHOENIX 2014T; however, we observe correct signs when indexing the frame at which the corresponding attention vector is maximised.
This suggests that alignment emerges from the attention
mechanism also for a full translation system.

\subsection{Discussion}
\label{subsec:discussion}
From our investigations in this work, we believe there are important and challenging
problems to be solved before
achieving large-vocabulary sign language \mbox{\textit{translation}}
from videos to spoken language.
First, significantly expanding the
coverage of the \textit{vocabulary}
of both languages is necessary,
and the current state of the art only
covers about 3K spoken language and 1K
sign language vocabularies~\cite{camgoz2020sign}. In preliminary experiments, we found
that a direct application of~\cite{camgoz2020sign} to
translation on the significantly broader vocabulary of 40K contained within
the subtitles of BSL-1K failed to converge to meaningful
results (for more details see appendix,
\if\sepappendix1{Sec.~C.1).}
\else{Sec.~\ref{subsec:app:processing}).}
\fi
In this work,
we have extended to an 11K spoken language
vocabulary, but the NLP literature typically
works with much larger vocabularies (e.g. a few hundred thousand words~\cite{dai2019transformer}).
Our attempts to move to 40K words did
not obtain sufficient-quality results.
Second, the \textit{alignment}
between text and video is far from
perfect in large-scale sign language datasets
which inserts significant amount of noise
in training. Our automatic alignment
attempts in this work did not obtain improvements.
Relying on sparse annotations for approximate
alignments limits the amount of data.
Third, most of the works,
including ours, focus on \textit{interpreted}
data, which has certain biases. In fact, the act of interpreting can cause a simplification in signing style and vocabulary, and even lead to a reduction in speed for comprehension ~\cite{bragg2019}. Datasets of native signers should be built to train strong, robust models that generalise at scale and in the wild. Given these observations,
we believe that future work that specifically targets
translation systems will benefit from addressing 
these challenges.
We refer to the appendix
\if\sepappendix1{(Sec.~A)}
\else{(Sec.~\ref{sec:app:impact})}
\fi
for a discussion of broader impact. %

\section{Conclusions}
\label{sec:conclusion}

We have presented an approach to localise signs in continuous sign language videos 
with weakly-supervised subtitles by leveraging the attention mechanism of a 
Transformer model trained on a video-to-text sequence prediction task.
We find that state-of-the-art translation models
have very low recall %
on a large-vocabulary dataset, but a satisfactory localisation accuracy %
through attention that allows us to annotate sign timings.
We automatically 
annotate hundreds of thousands of new signing instances through our learned 
attention and validate their quality by using them to train a sign language 
recognition model that surpasses the state of the art on the BSL-1K benchmark as 
well as a more robust sign language benchmark which is 18 times larger.
Future work can leverage our automatic annotations and recognition model 
for large-vocabulary sign language translation.

\noindent\textbf{Acknowledgements.}
This work was supported by EPSRC grant ExTol and a Royal Society Research Professorship.
We thank
Cihan~Camg\"{o}z, %
Himel~Chowdhury and Abhishek~Dutta for their help.

\balance
{\small
\bibliographystyle{ieee}
\bibliography{shortstrings,references,vgg_local,vgg_other}
}

\clearpage

\bigskip
{\noindent \large \bf {APPENDIX}}\\
\renewcommand{\thefigure}{A.\arabic{figure}}
\setcounter{figure}{0} 
\renewcommand{\thetable}{A.\arabic{table}}
\setcounter{table}{0} 

\appendix

This appendix to the main paper provides
a discussion on broader impact (Sec.~\ref{sec:app:impact}),
additional qualitative 
(Sec.~\ref{sec:app:qualitative})
and quantitative results
(Sec.~\ref{sec:app:exp}),
as well as implementation details
(Sec.~\ref{sec:app:details}).

\section{Broader impact}\label{sec:app:impact}

At present, computer vision technology to usefully assist signers remains in its infancy.  In large part, this stems from the high difficulty of achieving robust machine comprehension of sign language, which falls a long way short of human performance~\cite{bragg2019}.  
Our work, which focuses specifically on \textit{sign localisation}, takes steps towards enabling several practical applications that may become viable even when a full automatic understanding of sign language remains incomplete. These include: (1) \textit{sign language dictionary construction} to assist students who wish to learn sign language, (2) \textit{index construction} for video corpora, allowing individuals to search videos by the content of their signing, (3) \textit{wake-word spotting} for signing users of smart assistants like Alexa and Siri, (4) \textit{tools for linguists} to assist in the efficient analysis of existing signing data and (5) \textit{automatic large-scale dataset construction} to facilitate future research towards technology that will ultimately be able to provide useful products and services to the Deaf community.

The development of automatic, accurate sign localisation also has risks. Notably, It has the potential to be used for surveillance. Moreover, as with many computer vision methods employing deep neural networks (as ours does), the model is prone to fitting the training distribution closely.  As a result, it will be vital that products and services employing this technology ensure that their users are well-represented in the training data to avoid a disparity of performance across groups.

\section{Qualitative results}\label{sec:app:qualitative}
We refer to our supplementary video 
at our project webpage %
for additional visual results and illustrations.
First, we visualise the attention scores for sample test videos,
similarly to
\if\sepappendix1{Fig.~4}
\else{Fig.~\ref{fig:qualitative}}
\fi
of the main paper. To make it easier
to assess the localisation quality visually, we show
an example dictionary video corresponding to the
localised sign.
Next, we demonstrate the capability to
temporally localise signs in long continuous videos
using this attention mechanism on a training sequence.
Finally, we present our automatic annotations on the training set
(that we obtain through checking against subtitles),
which we use for sign language recognition training.
When grouping videos corresponding to the same word,
we observe a temporal alignment across samples.

\section{Additional experimental results}\label{sec:app:exp}
We provide additional quantitative analysis
through experimentation with different subtitle preprocessing
approaches (Sec.~\ref{subsec:app:processing}),
a detailed breakdown of performance for methods
incorporating sparse annotations
(Sec.~\ref{subsec:app:sparse}),
additional decoding strategies
to mine training examples (Sec.~\ref{subsec:app:decoding}),
and a recognition architecture study (Sec.~\ref{subsec:app:architecture}).

\subsection{Subtitle processing}
\label{subsec:app:processing}

All experiments in this section are reported on \testloc{}
to evaluate the Transformer training for the sequence prediction task.

\noindent\textbf{Stemming.} We experiment with stemming versus using the original subtitle words in Tab.~\ref{tab:app:stemming}.
The 11K annotation stems correspond to a vocabulary of 16K words.
We train a model by filtering the subtitles to these 16K words
without any further processing.
We observe no significant differences between the two models.
Note that, for a fair comparison, we stem the words at
evaluation time.

\begin{table}[h!]
    \setlength{\tabcolsep}{8pt}
    \centering
    \resizebox{0.99\linewidth}{!}{
        \begin{tabular}{lcccrr}
            \toprule
             & vocab. & Recall & Prec. & Loc$_{GD}$ & Loc$_{TF}$ \\
            \midrule
            Stemming & 11K & 16.5 & 37.2 & 66.1 & 44.5 \\
            No stemming & 16K & 16.0 & 35.5 & 66.8 & 52.3 \\ %
            \bottomrule
        \end{tabular}
    }
    \vspace{3pt}
    \caption{
        \textbf{Stemming subtitles:} We find that stemming
        might not be necessary for training. Note that
        for both models, we stem the words at evaluation.
    }
    \label{tab:app:stemming}
\end{table}

\begin{table}
    \setlength{\tabcolsep}{3pt}
    \centering
    \resizebox{0.99\linewidth}{!}{
        \begin{tabular}{lclc|ccrr}
            \toprule
            \% of subtitle & train & \multicolumn{1}{c}{test} & \#test\\
             vocabulary & vocab. & \multicolumn{1}{c}{vocab.} & subtitles & Recall & Prec. & Loc$_{GD}$ & Loc$_{TF}$ \\
            \midrule
            100\%  &26K  & 26K & 7588 & 14.9 & 35.6 & 67.9 & 49.9 \\
             & & 11K(11K) & 7497 & 15.9 & 36.0 & 68.0 & 50.2 \\
            \midrule
            75\% &19K  & 19K & 7567 & 15.2 & 36.0 & 67.7 & 38.9 \\
            & & 11K(10K) & 7497 &  16.2 & 36.3 & 67.8 & 39.6 \\
            \midrule
            50\% & 13K& 13K & 7516 & 15.9 & 36.8 & 66.6 & 52.4 \\
            &  & 11K(9K) & 7497 &  16.7 & 36.9 & 66.7 & 52.3 \\
            \midrule
            25\% &6K & 6K & 7271 & 17.0 & 40.0 & 66.3 & 52.0 \\
            &  & 11K(6K) & 7497 &  17.0 & 39.0 & 66.6 & 51.8 \\
            \midrule
            \midrule
            Annot. vocab. & 11K & 11K &7497 & 16.5 & 37.2 & 66.1 & 44.5 \\
            \bottomrule
        \end{tabular}
    }
    \vspace{3pt}
    \caption{
        \textbf{Vocabulary size:}
        We systematically
        change the training vocabulary of stems by
        taking subsets of the full subtitle vocabulary.
        We take the top 25\%, 50\%, 75\%, 100\% of stems according to
        their frequencies in the subtitles. Each trained model
        is tested twice (two rows per model): with
        (a) the same vocabulary used for training,
        (b) the comparable 11K vocabulary used in the
        rest of the experiments.
        Note that in (b),
        there might not be a full
        overlap between the train and test; the numbers in parenthesis
        represent the intersection. 
    }
    \label{tab:app:vocab}
\end{table}

\noindent\textbf{Vocabulary.} In this work,
we have used a vocabulary of 11K stems which is
determined based on the annotations.
In Tab.~\ref{tab:app:vocab}, we train
additional Transformer models by using vocabularies
determined by the subtitles. We sort the stems
appearing in all subtitles based on their frequencies.
We train with top 25\%, 50\%, 75\%, 100\% of all stems.
We observe that the models are not very sensitive
to the choice of training vocabulary.
Note that in all cases, we filter out the stop words
which do not have sign correspondences.

\begin{table}[t]
    \setlength{\tabcolsep}{8pt}
    \centering
    \resizebox{0.99\linewidth}{!}{
        \begin{tabular}{lcccrr}
            \toprule
            & vocab. & Recall & Prec. & Loc$_{GD}$ & Loc$_{TF}$ \\
            \midrule
            Without stop words & 11K & 16.5 & 37.2 & 66.1 & 44.5 \\ %
            With stop words & 11K & 13.9 & 25.9 & 69.5 & 52.5 \\ %
            \bottomrule
        \end{tabular}
    }
    \vspace{3pt}
    \caption{
        \textbf{Removing stop words:} We train a model by including
        the stop words (although these rarely have corresponding signs), and obtain lower performance (13.9\% recall).
    }
    \vspace{-5pt}
    \label{tab:app:stopwords}
\end{table}

\begin{table}[h!]
    \setlength{\tabcolsep}{8pt}
    \centering
    \resizebox{0.99\linewidth}{!}{
        \begin{tabular}{llc|ccrr}
            \toprule
            train & test & \#test\\
            vocab. & vocab. & subtitles & Recall & Prec. & Loc$_{GD}$ & Loc$_{TF}$ \\
            \midrule
            40K & 40K & 7413 & 9.5 & 7.5 & 70.3 & 27.1 \\
            40K & 16K & 7299 & 10.3 & 7.5 & 70.3 & 27.0 \\ %
            \bottomrule
        \end{tabular}
    }
    \vspace{3pt}
    \caption{
        \textbf{Naive translation with 40K vocabulary:} We
        report the results of training a model without stemming
        and without vocabulary filtering (except we filter to the 260K
        English vocabulary of Transformer-XL to remove the noise
        in the subtitles, due to OCR mistakes etc.).
        We test the model on (a) the same 40K vocabulary used for training,
        and (b) the 16K subset covering the annotations.
        Overall, we observe poor precision and recall.
    }
    \vspace{-5pt}
    \label{tab:app:translation}
\end{table}

\noindent\textbf{Stop words.}
In Tab.~\ref{tab:app:stopwords},
we train one model by keeping the stop words
and compare against our model. Note that
we determine the list of stop words according to
English stop words in the
\texttt{nltk.corpus}.
Qualitatively,
we observe frequent occurrence of the words ``and'' and ``to''
in the predictions.
The precision and recall metrics
reflect the reduced quality of the outputs as well.
Therefore, we filter out the stop words in all other
models.

\noindent\textbf{Naive translation.}
Tab.~\ref{tab:app:translation} reports
results for training a model with a large
vocabulary, without filtering and
without stemming. We again stem the words
at evaluation time. We observe poor performance
and highlight the difficulty of the translation
problem on in-the-wild sign language data.
A few qualitative predictions are provided below.
We note that while some examples have overlap
between ground truth and prediction (\#1, \#2, \#3),
many examples repeat the same prediction (\#4, \#5),
or output frequent words (\#6). As argued
in the discussion section of the main paper
\if\sepappendix1{(Sec.~4.6),}
\else{(Sec.~\ref{subsec:discussion}),}
\fi
we believe that the video-text alignment and large-vocabulary
sign recognition problems should become more advanced to
achieve in-the-wild translation.

\begin{Verbatim}[fontsize=\scriptsize, frame=single, framerule=0.1mm, commandchars=\\\{\}]
Example #1
Reference:  through your own admission your last
time in the \textbf{competition}
\textcolor{MidnightBlue}{Hypothesis: the \textbf{competition} is a competition}

Example #2
Reference:  lots of \textbf{water} to help digest such a meal
\textcolor{MidnightBlue}{Hypothesis: and then the \textbf{water} is the water}

Example #3
Reference:  \textbf{people} talk you see
\textcolor{MidnightBlue}{Hypothesis: i think the \textbf{people} were a good}

Example #4
Reference:  and just tease out the dead growth
\textcolor{MidnightBlue}{Hypothesis: and the whole thing is a little bit}

Example #5
Reference:  and how little we knew about the species
\textcolor{MidnightBlue}{Hypothesis: and the whole thing}

Example #6
Reference:  wrong here again going to give it how many
\textcolor{MidnightBlue}{Hypothesis: i think that is a good}
\end{Verbatim}

\subsection{Incorporating sparse annotations}
\label{subsec:app:sparse}

As in Sec.~\ref{subsec:app:processing}, all experiments in this section are reported on \testloc{}.

\noindent\textbf{Alignment loss on sparse annotations.}
Tab.~\ref{tab:app:alignmentloss} presents detailed
results on the incorporation of the alignment loss
as described in
\if\sepappendix1{Sec.~4.3}
\else{Sec.~\ref{subsec:mining}}
\fi
of the main paper.
Although minor improvements are observed with
the addition of such a loss term, we do not use
it in the final model for simplicity.

\begin{table}[h!]
    \setlength{\tabcolsep}{4pt}
    \centering
    \resizebox{0.99\linewidth}{!}{
        \begin{tabular}{lccc|rr|rr}
            \toprule
            & && & \multicolumn{2}{|c}{Loc. Acc. (GD)} & \multicolumn{2}{|c}{Loc. Acc. (TF)} \\
            $\lambda_{\mathcal{L}_{align}}$ & L& Recall & Prec. & layer 1/2 & [avg] &   layer 1/2 & [avg] \\
            \midrule
            0 & - & 16.5 & 37.2 &63.9/57.8 & [66.1] & 51.1/37.6 & [44.5] \\
            \midrule
            10 & avg & \textbf{16.8} & 37.5 & 64.7/59.2 & [66.0] & 51.4/36.1  & [45.2] \\
            100 & avg & 16.4 & 37.2 & 67.4/60.7 &  [68.0] & 52.8/42.3 & [48.0] \\
            1000&  avg & 14.4 & 34.5 & 68.9/59.0 & [67.3] &  52.5/55.8 & [56.6]  \\
            \midrule
            10 & 1 & \textbf{16.8} & 38.3 & 62.9/63.8 & [66.4] & 51.2/40.7 & [43.1] \\
            100 & 1 & 16.7 & 37.3 & 67.5/63.6 & [66.7] & 52.9/38.5 & [42.1] \\
            1000 & 1 & 15.7 & 33.5 & 59.4/69.8 & [69.6] & 57.0/35.6 & [48.7] \\
            \midrule
            10 & 2 & \textbf{16.8} & 37.4 & 65.8/59.2 & [67.3] & 51.7/36.8 & [47.1] \\
            100 & 2 & 16.2 & 37.7 & \textbf{68.5}/57.2 & [68.0] & 46.0/50.4 & [47.5] \\
            1000 & 2 & 14.8 & 35.5 & 67.1/59.2 & [66.3] & 42.5/\textbf{58.7} & [53.6] \\
            \bottomrule
        \end{tabular}
    }
    \vspace{3pt}
    \caption{
        \textbf{Alignment loss on sparse annotations:}
        We experiment with different weighting terms for
        the alignment loss ($\lambda_{\mathcal{L}_{align}}$)
        in addition to the classification loss during
        subtitle training. We define the loss on various
        attention layers (L) of a 2-layer architecture.
        We observe minor improvements.
    }
    \label{tab:app:alignmentloss}
\end{table}

\noindent\textbf{Curriculum learning with sparse annotations.}
Tab.~\ref{tab:app:curriculum} reports
results with and without the curriculum strategy
described in
\if\sepappendix1{Sec.~4.3}
\else{Sec.~\ref{subsec:mining}}
\fi
of the main paper.
We obtain minor improvements with pretraining
the Transformer on shorter temporal segments containing
only 1 annotated sign, finetuning the model later on 2 and 3 signs.
Note that this model uses 1 layer in both encoder
and decoder unlike other experiments which use 2 layers (we note from our Transformer layer ablations reported the main paper that this does not dramatically affect localisation performance).

\begin{table}[h!]
    \setlength{\tabcolsep}{3pt}
    \centering
    \resizebox{0.99\linewidth}{!}{
        \begin{tabular}{lccrr}
            \toprule
            Training schedule & Recall & Prec. & Loc$_{GD}$ & Loc$_{TF}$ \\
            \midrule
            No curriculum: Subtitle & 15.8 & 36.4  & 65.9 & 44.8 \\
            With curriculum: 1$\rightarrow$2$\rightarrow$3$\rightarrow$Subtitle & 16.0 & 37.1  & 66.6 & 44.3 \\
            \bottomrule
        \end{tabular}
    }
    \vspace{3pt}
    \caption{
        \textbf{Curriculum learning with sparse annotations:}
        We observe minor improvements by incorporating curriculum
        learning, which gradually extends the temporal window
        of the input video.
        Note that a 1-layer encoder-decoder architecture is used
        for this experiment.
    }
    \label{tab:app:curriculum}
\end{table}

\begin{table}[th!]
    \setlength{\tabcolsep}{3pt}
    \centering
    \resizebox{0.99\linewidth}{!}{
        \begin{tabular}{lccrr}
            \toprule
            Training subtitles & Recall & Prec. & Loc$_{GD}$ & Loc$_{TF}$ \\
            \midrule
            662K not aligned & 6.8 & 16.3 & 67.1 & 27.0 \\
            183K not aligned (subset) & 6.2 &  15.0 & 65.4 & 25.5 \\ %
            230K aligned & 15.4 & 38.3 & 66.7 & 51.5 \\
            301K coarse (pad $\pm$2-sec) & 13.9 & 45.2 & 67.6 & 48.3 \\
            \midrule
            183K coarse & 16.5 & 37.2 & 66.1 & 44.5 \\
            \bottomrule
        \end{tabular}
    }
    \vspace{3pt}
    \caption{
        \textbf{Subtitle-video alignment:} Our coarse alignment,
        which uses the assumption that the subtitles that have at
        least one annotation within the subtitle timestamp is aligned
        with its video, obtains the best performance over other
        alignment variants or using no alignment.
    }
    \vspace{-8pt}
    \label{tab:app:subalign}
\end{table}

\noindent\textbf{Video-subtitle alignment.} 
Tab.~\ref{tab:app:subalign} details our experiments
which highlight the importance of video-subtitle
alignment. When using all subtitles without considering
whether they contain an annotation or not (662K subtitles),
we obtain poor recall on the test set where there is at least
one high-confidence ($\ge$0.9) annotation. To keep the number
of training subtitles same as our final model, we also experiment
with taking a random subset of 183K subtitles, and observe
a similar outcome of poor performance.
When using active signer detection
and sparse annotations to apply a simple
algorithm to align the subtitles, we get to 230K
training subtitles that have at least 1 annotation; however,
this model does not impact the results significantly.
To take the uncertainty into account, we also experiment with
padding
$\pm$2 seconds at the start and end of the subtitle
times to input more video features to the model.
However, this model also reduces the recall.
Our simple coarse alignment strategy of
using subtitles that have at least 1 annotation
results in the best performance.

\subsection{Decoding strategies}
\label{subsec:app:decoding}

In
\if\sepappendix1{Sec.~4.3}
\else{Sec.~\ref{subsec:mining}}
\fi
of the main paper,
we have described different decoding mechanisms
to mine new training annotations by applying
the Transformer model. Here, we provide two
more strategies to complement
\if\sepappendix1{Tab.~3}
\else{Tab.~\ref{tab:yield}}
\fi
(a) \textbf{1 best:} choosing the hypothesis with
the highest recall when applying beam search
with size 10, (b) \textbf{TF prediction:} decoding with teacher forcing and forming a hypothesis using the model's prediction at every step, then filtering the hypothesis by only keeping the tokens that are also present in the corresponding subtitle (same as with GD); this is an alternative form of the TF baseline -- here we also use the attentions of the first layer only.
The new results are denoted with bold font
in Tab.~\ref{tab:app:decoding} for \testrecnew{}.
The best result is achieved by simple
greedy decoding (GD) which has
smaller but more noise-free sign
localisations.

\begin{figure}[h]
    \centering
    \includegraphics[width=0.47\textwidth]{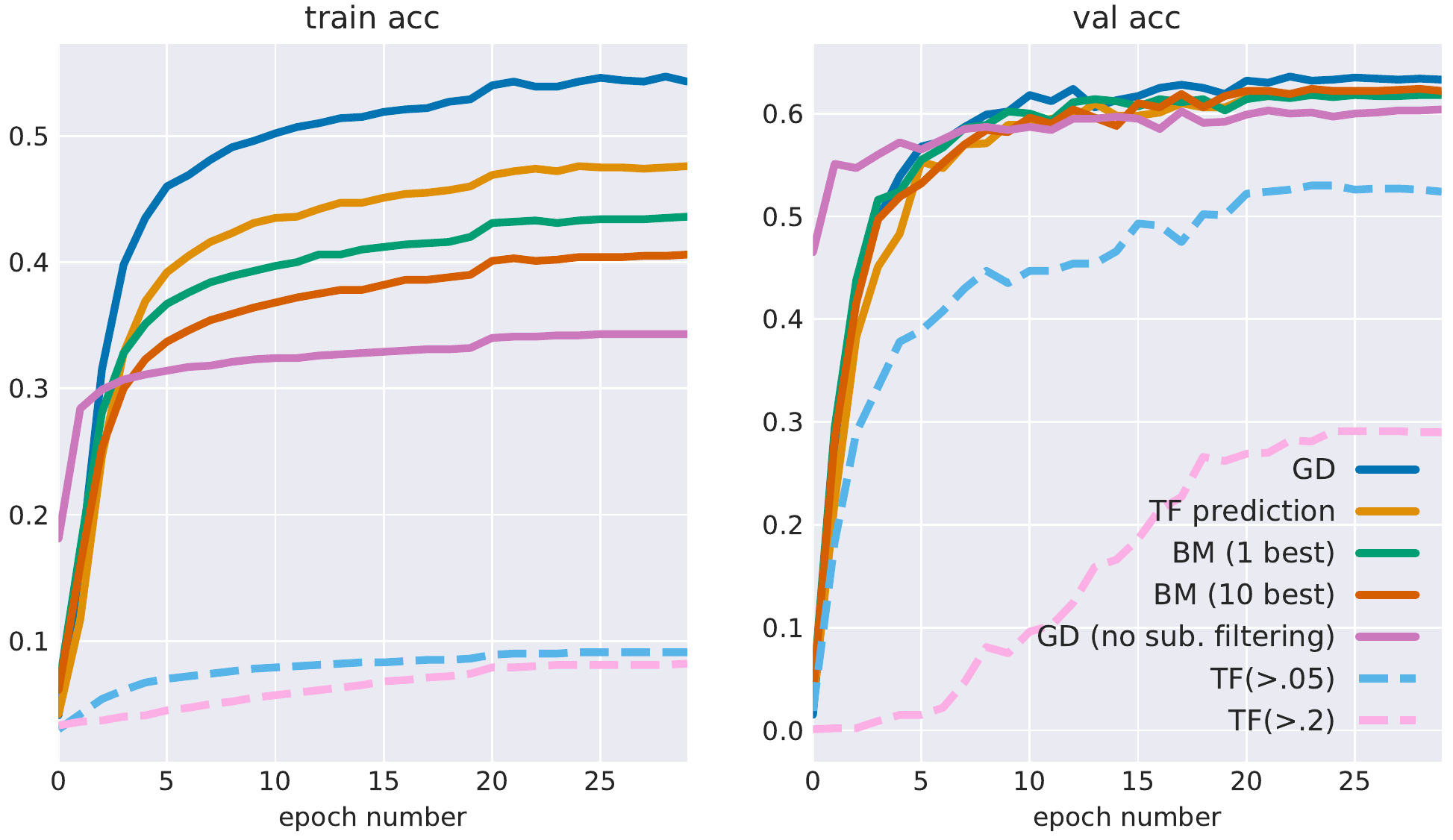}
    \caption{\textbf{
            Training recognition with attention spottings:} 
        We plot the training (left) and validation (right) accuracy curves
        against the epoch number for different MLP models,
        corresponding to different decoding strategies
        to mine training examples. The legend corresponds to descriptions
        in Tab.~\ref{tab:app:decoding} and
        \if\sepappendix1{Tab.~3}
        \else{Tab.~\ref{tab:yield}}
        \fi
        of the main paper. See Sec.~\ref{subsec:app:decoding} and
        \if\sepappendix1{Sec.~4.3}
        \else{Sec.~\ref{subsec:mining}}
        \fi
        of the main paper
        for details.
        We conclude that increased noise in teacher forcing
        mechanism (dashed), despite its large yield, makes learning
        difficult.
    }
    \vspace{-8pt}
    \label{fig:app:decoding}
\end{figure}

Fig.~\ref{fig:app:decoding} shows several training plots
corresponding to different decoding mechanisms.
The curves suggest that mining more examples
with higher noise results in low training performance.
The plotted metric is top-1 per instance accuracy
over 30 training epochs.

\subsection{Recognition architecture study}
\label{subsec:app:architecture}

We present an experimental study
for the architecture design
of our MLP which is used for recognition.
Tab.~\ref{tab:app:architecture}
summarises the results on \testrecnew{} for training
with all M+D+A annotations, i.e.,
our best model in
\if\sepappendix1{Tab.~4}
\else{Tab.~\ref{tab:sota}}
\fi
of the main paper.
While the results are not significantly
different, we observe minor improvements
with increased capacity, which quickly saturates
when adding more layers.

\begin{table}[th!]
    \setlength{\tabcolsep}{8pt}
    \centering
    \resizebox{0.99\linewidth}{!}{
        \begin{tabular}{lrrr|rr}
            \toprule
            & \#subtitles  & \#ann. & \#ann. & top-1   & top-1 \\
            Spotting mode    & unannot.   &  11K &   1K &         per-inst   & per-cls\\
            \midrule
            TF ($\geq .05$)   & 457K & 2.3M & 754K &  38.7  & 14.4  \\
            BS (10 best)        & 109K  & 329K & 166K & 49.6 & 22.7\\
            \textbf{BS (1 best)} & 109K & 316K & 161K & 50.7 & 23.3 \\
            \textbf{TF prediction} & 57K & 195K & 110K & 51.3 & 22.5 \\  %
            GD                  & 53K  & 188K & 107K & 53.9 & 24.7 \\
            \bottomrule
        \end{tabular}
    }
    \vspace{3pt}
    \caption{
        \textbf{Other decoding strategies:}
        TF: teacher-forced decoding, filtering with a threshold on attention scores; BS (10 best): beam search decoding with beam size 10 - all returned hypotheses are used when looking for new instances; BS (1 best): the same beam search is performed (with beam size 10) but only the hypothesis with the highest recall is used; TF prediction: teacher-forced decoding, using the hypothesis predicted by the model and checking against subtitle;
        GD: greedy decoding.
        The strategies shown in bold font refer to experiments not included in the main paper and are described in more detail in Sec.~\ref{subsec:app:decoding}. For the rest, we refer to
        \if\sepappendix1{Sec.~4.3}
        \else{Sec.~\ref{subsec:mining}}
        \fi
        of the main paper. 
    }
    \vspace{-8pt}
    \label{tab:app:decoding}
\end{table}

\begin{table}[h!]
    \setlength{\tabcolsep}{4pt}
    \centering
    \resizebox{0.95\linewidth}{!}{
        \begin{tabular}{lcccc}
            \toprule
            & \multicolumn{2}{c}{per-instance} & \multicolumn{2}{c}{per-class} \\
            Architecture & top-1 & top-5 & top-1 & top-5  \\
            \midrule
            1024$\rightarrow$res(1024)$\rightarrow$512$\rightarrow$256 & 65.1 & 82.6 & 38.0 & 56.4 \\
            \midrule
            1024$\rightarrow$512$\rightarrow$256 & 64.6 & 82.3 & 36.5 & 55.1 \\
            1024$\rightarrow$256 & 63.8 & 82.0 & 35.2 & 53.9 \\
            \midrule
            1024$\rightarrow$res(1024)$\rightarrow$512$\rightarrow$128 & 64.9 & 82.6 & 37.4 & 55.9 \\
            1024$\rightarrow$res(1024)$\rightarrow$512$\rightarrow$512 & 65.3 & 82.6 & 38.4 & 57.0 \\
            1024$\rightarrow$res(1024)$\rightarrow$512$\rightarrow$1024 & \textbf{65.5} & \textbf{82.9} & \textbf{39.0} & \textbf{57.4} \\
            \midrule
            1024$\rightarrow$res(1024)$\rightarrow$512 & 65.3 & \textbf{82.9} & 37.9 & 56.6 \\
            1024$\rightarrow$res(1024)$\rightarrow$512$\rightarrow$512$\rightarrow$1024 & \textbf{65.5} & 82.7 & \textbf{39.0} & \textbf{57.4} \\
            1024$\rightarrow$res(1024)$\rightarrow$512$\rightarrow$512$\rightarrow$512$\rightarrow$1024 & 65.0 & 82.4 & 38.3 & 57.1  \\
            \bottomrule
        \end{tabular}
    }
    \vspace{3pt}
    \caption{
        \textbf{Architecture study for recognition:} We experiment with
        different number of layers for the sign recognition model.
        The input dimensionality is 1024,
        which is a temporally-averaged I3D embedding over 16 frames. 
        The output is 1064-dimensional class probabilities.
        The top row is what is reported in the main paper
        (corresponding to Fig.~\ref{fig:app:mlp}).
        We observe minor improvements by increasing the network
        capacity.
    }
    \label{tab:app:architecture}
\end{table}

\section{Implementation details}\label{sec:app:details}

\subsection{Application of M \cite{Albanie20} and D \cite{Momeni20}}
\label{subsec:app:md}

As explained in
\if\sepappendix1{Sec.~4.1,}
\else{Sec.~\ref{subsec:datasets},}
\fi
we apply the method of \cite{Albanie20} to localise
signs through mouthing cues on a large vocabulary
of words beyond 1K (which is used in the original work).
In particular, we query 36K words, and out of these,
a vocabulary of 15K words are localised with confidence above 0.7.
When applying the method of \cite{Momeni20} to localise signs through
similarity matching with dictionary videos,
we query 9K signs from the full BSLDict dataset
with search windows of $\pm$4 seconds padding around the subtitle timestamps.
The resulting sign localisations with confidence above 0.7
cover a vocabulary of 4K words. The combination of
these two methods gives us a total vocabulary of 16K words, which results in the 11K stems 
used for our Transformer training.

\begin{figure}
    \centering
    \includegraphics[width=0.47\textwidth]{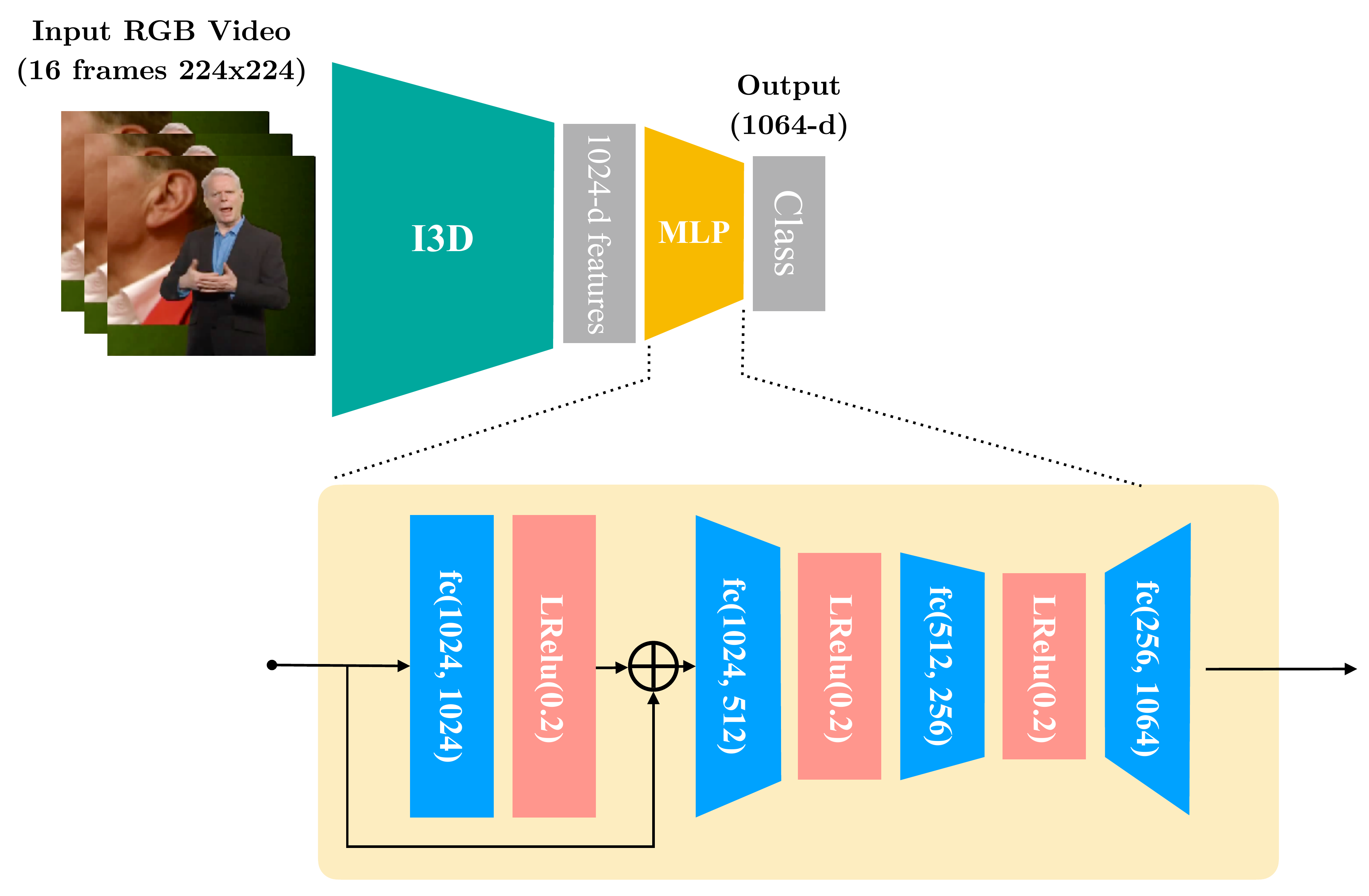}
    \caption{\textbf{MLP architecture for recognition:} 
        We follow \cite{momeni20watchread},
        and use one residual connection,
        followed by 3 more fully connected layers
        on top of the I3D pre-extracted features. 
    }
    \label{fig:app:mlp}
\end{figure}

\subsection{Architecture and training details}

Our Transformer model consists of 2 attention layers for
both encoding and decoding. The input 1024-dimensional
video feature is mapped to 512 dimensions with a linear layer.
Then 512-d embeddings are used both for output words and
input videos. We use 2 heads in each attention layer.

When reporting localisation accuracy, we average the
encoder-decoder attention scores over the 2 heads.
We take the first layer attention for teacher forcing (TF)
and the average over two layers for greedy decoding (GD).
We mark a correct localisation if the maximum location
over the input video is within $\pm$2 feature frames from
the annotation time. This is because one sign approximately
lasts for 7-13 frames (at 25fps)~\cite{momeni20watchread}
and our features are extracted with a stride of 4 frames,
making our valid window duration $\pm$8 video frames.
This also accounts for some uncertainty in the `ground-truth'
annotation times which are obtained automatically.

We detail our MLP architecture in Fig~\ref{fig:app:mlp}.
We use a design similar to~\cite{momeni20watchread}.
The architecture study in Sec.~\ref{subsec:app:architecture}
reports variations of this model.
We train it for 30 epochs, with an initial learning
rate of $1e^{-2}$ reduced by a factor 10 at the 20th and 25th epoch.

\subsection{Infrastructure} \label{subsec:app:infra}
We use Nvidia M40 graphics cards for our experiments.
The video-subtitle Transformer model trains in 10 hours on a single GPU.
The annotation mining time is roughly 30 minutes to obtain 107K
annotations, i.e., Transformer forward pass runtime over 1302h of training
videos (duration of 685K subtitles padded with $\pm$2 seconds) on a single GPU.
The final best MLP (M+D+A) for sign recognition trains in 7 hours on a single GPU.
The M+D I3D backbone is trained with 4 GPUs over a duration of 1 week.

\end{document}